\documentclass{article}

\usepackage{graphicx} % more modern
\usepackage{subfigure} 

\usepackage{natbib}

\usepackage{algorithm}
\usepackage{algorithmic}

\usepackage{hyperref}

\usepackage[accepted]{icml2013}
\icmltitlerunning{Horizontal and Vertical Ensemble with Deep Representation for Classification}

\begin{document} 

\twocolumn[
\icmltitle{Horizontal and Vertical Ensemble with Deep Representation for Classification}

% unless you've provided the [accepted] option to the icml2013
\icmlauthor{Jingjing Xie}{xiejingjing113@gmail.com}

\icmlauthor{Bing Xu}{antinucleon@gmail.com}

\icmlauthor{Zhang Chuang}{zhangchuang@bupt.edu.cn}
\icmladdress{Beijing University of Posts and Telecommunications,
            10th Xitucheng Rd., Beijing, China, 100876}
\icmlkeywords{boring formatting information, machine learning, ICML}

\vskip 0.3in
]

\begin{abstract} 
Representation learning, especially which by using deep learning, has been widely applied in classification. However, how to use limited size of labeled data to achieve good classification performance with deep neural network, and how can the learned features further improve classification remain indefinite. In this paper, we propose Horizontal Voting Vertical Voting and Horizontal Stacked Ensemble methods to improve the classification performance of deep neural networks. In the ICML 2013 Black Box Challenge, via using these methods independently, Bing Xu achieved 3rd in public leaderboard, and 7th in private leaderboard; Jingjing Xie achieved 4th in public leaderboard, and 5th in private leaderboard.

\end{abstract} 

\section{Introduction}
Classification is one of the most important machine learning tasks. Besides classification algorithms, the performance of classifier is heavily dependent on the set of data representations on which they are applied. Traditionally, data representations are hand-crafted, with prior knowledge or hypotheses of the human designers. Then the classifiers with the designed representations (or features) are trained by fitting the labeled data, expected to give a good class prediction on test data inputs.

However, the increasing size of data in real world and the variety of learning tasks bring challenges to this traditional paradigm. Practically, labeled data is rare, but unlabeled data is always abundant. Although there are some less expensive ways to obtain labels, automatically learning representations from data would be more efficient. Furthermore, it has been proved that in some fields automatic representation learning can work better, even if human feature engineering is still powerful. In the ICML 2013 Black Box Challenge\footnote{http://www.kaggle.com/c/challenges-in-representation-learning-the-black-box-learning-challenge}, both labeled and unlabeled data are provided to players without prior knowledge about what the data really is. So we resort to deep learning and design a deep neural network which consists 5 layers of denoising auto-encoder and 3 maxout layers, with more than 16 million parameters in total. We use all the 130 thousand unlabeled data to pre-train the stacked denoising auto-encoders and fine-tune the huge deep network with only 1,000 training examples. As the task is data classification, it’s natural to ask: how to use so little labeled data to train a large deep network with robust classification result? Can the hierarchical representations in the deep architecture help improve the performance of classification?

In this work, we describe our method of training deep neural networks for classification with both labeled and unlabeled data. We also proposed three methods called Vertical Voting, Horizontal Voting and Horizontal Stacked Ensemble to improve the classification accuracy and robustness of deep network. Their performance and combination strategies are also discussed.

\section{Background} 
\begin{figure}[tbp]
\vskip 0.2in
\begin{center}
\centerline{\includegraphics[width=\columnwidth]{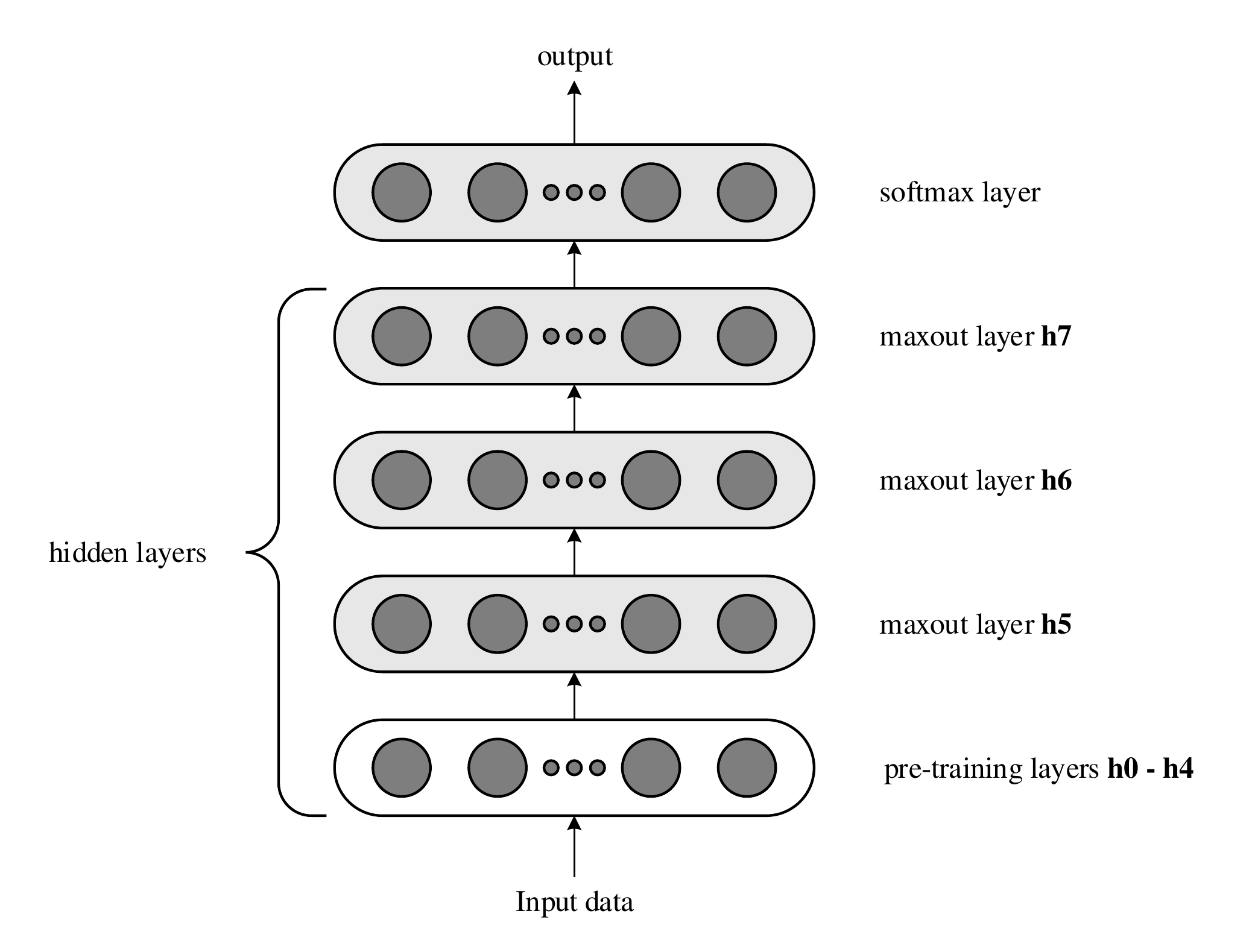}}
\caption{The architecture of deep neural networks. In this example, the deep network has 5 stacked auto-encoder layers (h0 - h4) which are represented by a single pre-training layer for simplicity. Upon them are 3 maxout layers (h5 - h7) and a softmax layer.}
\label{network_total}
\end{center}
\vskip -0.2in
\end{figure}

%Representation learning is to transform the data into different representations, and learn the effective mapping from input to expected output. 
A deep neural network applies combined transformations to input data, and produces representations with an increasing level of abstraction and complexity. The architecture of a deep neural network is drawn in Figure~\ref{network_total}. Input data are processed in a deep architecture of transformations, and generate desired output at the end. Usually, there are pre-training layers on the bottom of the architecture, which are built with Restricted Boltzmann Machine (RBM) \cite{smolensky1986information,hinton2006fast} or auto-encoder \cite{le1987modeles,bourlard1988auto,hinton1994autoencoders} layers. The layers above them such as sigmoid, tanh, and maxout \cite{goodfellow2013maxout} layers, together with pre-training layers, are collectively called hidden layers. At the top is the softmax layer which produces probabilities for each class as output.

The training of deep neural networks has two phases \cite{bengio2009learning}. The first phase is layer-wise unsupervised pre-training \cite{hinton2006fast,bengio2007greedy} which makes use of unlabeled data, adjusts the parameter of pre-training layers, and initializes the deep neural network to a data-dependent manifold \cite{erhan2009difficulty}. In the second phase, all parameters in the network are fine-tuned under the supervision of labeled data. The softmax layer on the top produces the probabilities of each class for each example. An alternative to obtain class prediction is to train a standard classifier (such as Random Forest or SVM) with learned data representations \cite{bengio2012representation}.

%An important property of the deep architecture is the representation hierarchy. Each layer of the network generates representation of input data. The representations in previous layer is taken by the next layer as its input. Through accumulated combination and transformation, representations become increasingly complicated and abstract, which lead to a hierarchy of representations. 

\section{Vertical Voting, Horizontal Voting and Horizontal Stacked Ensemble}

We proposed a series of methods to improve the performance of classification. These methods are Vertical Voting, Horizontal Voting and Horizontal Stacked Ensemble.

\subsection{Vertical Voting}
The softmax layer generates predictions by using the top level data representation. All the lower level representations are discarded. However, lower level representations of data may contribute to classification themselves. For example, word is a kind of low level data representation. Some words can be strongly indicative for a topic, but they may lost when deep neural network parsing the sentence into a high level representation. To help classification, we propose a method called Vertical Voting. This method ensembles a series of classifiers whose inputs are the representation of intermediate layers. A lower error rate is expected because these features seem diverse.

The procedure of Vertical Voting method is shown in Algorithm~\ref{alg:vdv}.

\begin{algorithm}[tbp]
   \caption{Vertical Voting}
   \label{alg:vdv}
\begin{algorithmic}
   \STATE {\bfseries Input:} training data $X$, test data $x$, target $y$, neural network $N$, max epoch $E$, objective epoch $e$, selected hidden layers $\Omega=\{\omega_1,\dots, \omega_n\}$, classification algorithm set $A$\\
   Initialize SGD trainer for $N$\\
   Initialize a list $Preds=[\ ]$\\
   Initialize $iteration=0$\\
   \REPEAT
   \STATE Use $X$ and $y$ to do one epoch back-propagation training on $N$\\
   $iteration = iteration + 1$
   \IF{$iteration=e$}
   \STATE Input $X$ and $x$ into $N$, get $X$ and $x$'s representation pairs set $R$ in each layer $\omega_i \in \Omega$: $R=\{(X_{\omega_1}, x_{\omega_1}),\dots, (X_{\omega_n}, x_{\omega_n})\}$\\
   \FOR{each pair $r_j\in R$}
   \STATE Train classifier $c$ using an algorithm $a\in A$, with training data $X_{\omega_j}$ and $y$\\
   Add $c$'s probabilistic prediction vector $p_{\omega_j}$ on $x_{\omega_j}$ to $Preds$
   \ENDFOR\\
   
   \ENDIF
   \UNTIL{$iteration > E$}\\
   $Pred = \sum_{p_{\omega_i}}^{p_{\omega_i} \in Preds}p_{\omega_i}$\\
   \STATE {\bfseries Output:} $argmax(pred)$
\end{algorithmic}
\end{algorithm}

\subsection{Horizontal Voting}
\begin{figure}[htbp]
\vskip 0.2in
\begin{center}
\centerline{\includegraphics[width=\columnwidth]{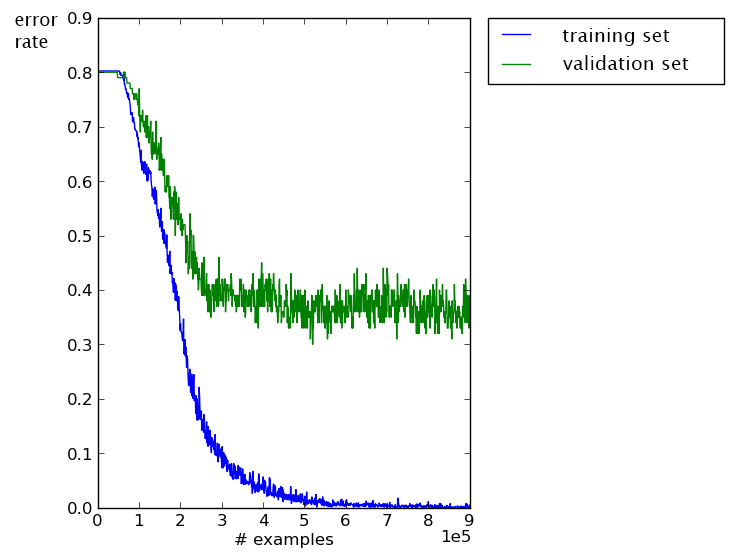}}
\caption{Learning curve of a deep network. 90\% of the training set is kept for training and 10\% is for validation.}
\label{m_curve}
\end{center}
\vskip -0.2in
\end{figure}

If appropriate network architecture and learning rate are chosen, the error rate of classification would first decline and then tend to be stable with the training epoch grows. But when size of labeled training set is too small, the error rate would oscillate, as shown in the validation set curve Figure~\ref{m_curve}. Although dropout \cite{hinton2012improving} helps a little, it is still overfit. So it is difficult to choose a ``magic" epoch to obtain a reliable output. To reduce the instability, we put forward a method called Horizontal Voting. First, networks trained for a relatively stable range of epoch are selected. The predictions of the probability of each label are produced by standard classifiers with top level representation of the selected epoch, and then averaged. 

The procedure of Horizontal Voting is shown in Algorithm~\ref{alg:hdv}.

\begin{algorithm}[htbp]
   \caption{Horizontal Voting}
   \label{alg:hdv}
\begin{algorithmic}
   \STATE {\bfseries Input:} training data $X$, test data $x$, target $y$, neural network $N$, max epoch $E$, selected epoch range $(L, H)$\\
   Initialize SGD trainer for $N$\\
   Initialize $iteration=0$\\
   Initialize a list $Preds=[\ ]$
   \REPEAT
   \STATE Use $X$ and $y$ to do one epoch back-propagation training on $N$\\
   $iteration = iteration + 1$
   \IF{$iteration>L$ and $iteration<H$}
   \STATE Put $x$ into $N$, get softmax output vector $pred_i$\\
   Add $pred_i$ to $Preds$\\
   \ENDIF\\
   \UNTIL{$iteration > E$}\\
   $Pred = \sum_{pred_i}^{pred_i \in Preds}pred_i$\\
   \STATE {\bfseries Output:} $argmax(pred)$
\end{algorithmic}
\end{algorithm}

\subsection{Horizontal Stacked Ensemble}
Sergey Yurgenson suggested a non-linear horizontal ensemble method\footnote{http://www.kaggle.com/c/challenges-in-representation-learning-the-black-box-learning-challenge/forums/t/4674/models?page=2} for shallow neural network, which has significantly improved the accuracy of classification. This method can be extended to deep neural networks. Similar to the horizontal voting method in section 3.2, it takes the output of networks within a continuous range of epoch. The following step is similar to Stacked Generalization method. All these outputs are collected to form a new feature space for classification.

The procedure of Horizontal Stacked Ensemble method is shown in Algorithm~\ref{alg:hne}.

\begin{algorithm}[tb]
   \caption{Horizontal Stacked Ensemble}
   \label{alg:hne}
\begin{algorithmic}
   \STATE {\bfseries Input:} training data $X$, test data $x$, target $y$, neural network $N$, max epoch $E$, selected epoch range $(L, H)$, classification algorithm set $A$\\
   Initialize SGD trainer for $N$\\
   Initialize $iteration=0$\\
   Initialize a list $Preds_x=[\ ]$\\
   Initialize a list $Preds_X=[\ ]$
   \REPEAT
   \STATE Use $X$ and $y$ to do one epoch back-propagation training on $N$\\
   $iteration = iteration + 1$
   \IF{$iteration>L$ and $iteration<H$}
   \STATE Put $x$ into $N$, get softmax output vector $pred_i$\\
   Add $pred_i$ to $Preds_x$\\
   Put $X$ into $N$, get softmax output vector $pred_i$\\
   Add $pred_i$ to $Preds_X$\\
   \ENDIF\\
   \UNTIL{$iteration > E$}\\
   Reshape $H-L-1$ softmax output vectors in $Pred_X$ to a single feature vector $F_X$ into a single vector of $(H-L-1) \times\ num\ of\ classes$ dimension\\
   Reshape $H-L-1$ softmax output vectors in $Pred_x$ to a single feature vector $F_x$ into a single vector of $(H-L-1) \times\ num\ of\ classes$ dimension\\
   Train classifier $c$ by using an algorithm $a\in A$, use training data $F_X$ and $y$\\
   Put $F_x$ into $c$, get final prediction $Pred$
   \STATE {\bfseries Output:} $argmax(pred)$
\end{algorithmic}
\end{algorithm}

\section{Models and Experiments}
The Black Box dataset provided by ICML 2013 Representation Learning Challenge is used. This dataset provides 1,000 labeled training examples, 10,000 test examples halved to public and private test, together with 135,735 unlabeled data for algorithms to exploit. The data is in 1,875 dimension, and is required to be divided to 9 classes. No prior knowledge can help since the data is human unreadable.

The deep neural network is chosen as the key for this challenge, and 6 models are designed. Model 1 is a single traditional shallow neural network and model 2 is a deep neural network. Model 2 is used as benchmark for the following models to test the efficiency of our methods, and explore the strategies of method combination. The models are described below. All of the experiments share same  learning rate and momentum. 

Most of the models are trained with open-source tools Theano \cite{bergstra2010theano}, PyLearn2 \cite{Pylearn2} and scikit-learn \cite{scikit-learn}.

\subsection{Model 1: Shallow neural network (without pre-training)}
Model 1 is a traditional shallow neural network without pre-training layers. The 1000 labeled data is the input of 3 maxout layers and a softmax layer. The number of neurons is 1875(input)-1500-1500-1500-9(output). 

\subsection{Model 2: Deep neural network with unsupervised pre-training}
Model 2(see Figure~\ref{network_total} for the architecture) add unsupervised pre-training layers to model 1. 5 denoising auto-encoder layers are used to take advantage of more than 130 thousand unlabeled data. The number of neurons is  1875(input)-1500-1000-1500-1200-1500-1500-1500-1500-9(output).

\subsection{Model 3: Deep neural network with Vertical Voting}
Model 3 adds Vertical Voting to Model 2 to test its effectiveness. Representations in the 3 maxout layers (h5-h7 in Figure~\ref{network_total}) vote for the prediction.

\subsection{Model 4: Deep neural network with Horizontal Voting}
In model 4, Horizontal Voting is introduced to the network in model 2. Deep neural networks that are trained from 651 to 850 epoch are averaged.

\subsection{Model 5: Deep neural network combined Horizontal and Vertical Voting}
Model 5 implements both Vertical Voting and Horizontal Voting based on Model 2. For every training epoch, Random Forest's prediction given by 3 hidden layers (h5-h7 in Figure~\ref{network_total}) are Vertically Voted. The process is repeated over the network of 651 to 850 epoch, then the 200 predictions are Horizontally combined.

\subsection{Model 6: Deep neural network with Horizontal Stacked Ensemble}
The model 6 introduces Horizontal Stacked Ensemble to Model 2. Also, 200 networks which are trained for 651 to 850 epoch are ensembled .

\section{Results and Discussion}
This section shows the result of each model, and discusses their performance. The gap in classification accuracy between model 1 and model 2 shows the contribution of pre-training layers. When compared with model 2, results of model 3 and model 4 may prove the effectiveness of the Vertical and Horizontal Voting respectively. In model 5 we can test the performance of both Voting methods. Model 6 and model 4 use different ensemble method and their performances can be compared.

\subsection{Model 1 and Model 2}
The classification of model 1 and model 2 are implemented with softmax function. The classification accuracy is shown in Table~\ref{m5}. Pre-training layers contribute to 20.19\% and 19.09\% score of model 2 in public and private test set respectively.

\subsection{Model 3}
Following the Vertical Voting method, Random Forest(with $n_{estimates} = 500$) provides predictions for representations generated by hierarchical layers (h5-h7) of the network. The classification accuracy of each prediction and the voted version are shown in the row 1-2 of Table~\ref{m31}. Note that the representation in h7 is classified by Random Forest here, while processed by softmax in model 2. The row 3-4 of Table~\ref{m31} shows the performance of Vertical Voting in another deep network, where the top maxout layer (h7) is replaced by a rectified linear layer.

By comparing columns of Table~\ref{m31}, it is observed that lower level representations do not play that well as higher level ones as we may had foreseen. However, the voted predictions in row 1-2 of Table~\ref{m31} do not have the highest accuracy, though the case in row 3-4 meet our expectation. We have three guesses that may be responsible for this phenomenon. The first guess is that representations in different layers of the same network do not provide good feature diversity. The second is that since overfitting exists in the deep network itself, ensembling a series of such models may deteriorate the performance. And the third is, adjusting the weight of voting may lead to better result. Validations of these guesses are beyond the scope of this paper.

\begin{table*}[htbp]
\caption{Classification accuracy of Random Forest for layer h5-h7, and the voted result. Row 1-2 is for model 3, and row 3-4 is for another deep neural network with Vertical Voting method. The best score of each experiment is in bold.}
\label{m31}
\vskip 0.15in
\begin{center}
\begin{small}
\begin{sc}
\begin{tabular}{l|llll}
    \hline
    ~                             & RF for h5 & RF for h6        & RF for h7        & Voted   \\ \hline
    Accuracy(public test set)     & 0.62440   & 0.66140          & \textbf{0.66240} & 0.65920 \\
    Accuracy(private test set)    & 0.62800   & \textbf{0.65760} & 0.65480          & 0.65620 \\ \hline
Accuracy(public test set)   & 0.63800   & 0.67220    & 0.67000 	 & \textbf{0.67240} \\
    Accuracy(private test set)  & 0.62760   & 0.65380    & 0.65720    & \textbf{0.65960} \\ \hline    
    \end{tabular}
\end{sc}
\end{small}
\end{center}
\vskip -0.1in
\end{table*}

\subsection{Model 4}
To obtain a smoother learning curve, we choose a learning rate $0.025$. Then following the Horizontal Voting method, a epoch range (650, 850] is chosen. The classification error rate statistic of the picked 200 examples is listed in Table~\ref{m_table}, which indicates a big oscillation. But by voting the prediction of 200 examples, the risk brought by a badly chosen epoch is greatly reduced.

As the Table~\ref{m5} shows, the result of horizontally voting achieved \textbf{0.68220} in public test set, and \textbf{0.67240} in private test set, which is the best score among our experiments during the challenge. Improvements are also achieved on networks of different structure. That shows horizontal Voting method can effectively produce a better and more robust performance of a deep neural network.

\begin{table}[htbp]
\caption{Classification error rate statistic of the 200 examples, calculated on validation set (10\% of the training set).}
\label{m_table}
\vskip 0.15in
\begin{center}
\begin{small}
\begin{sc}
 \begin{tabular}{llll}
    \hline
    Min         & Max         & Mean        & Standard Error\\ \hline
    0.309999   & 0.439999   & 0.375427   & 0.024364 \\ \hline
    \end{tabular}
\end{sc}
\end{small}
\end{center}
\vskip -0.1in
\end{table}

\subsection{Model 5}
Model 5 applies both Vertical and Horizontal Voting method to model 2. Table~\ref{m5} lists the classification accuracy of model 5, also make the performance of each model convenient to compare.

As we have observed in 5.3, the effectiveness of  Vertical Voting is not stable. The performance of model 5 is also influenced, if compared with model 4. In public test set the accuracy improves a little, but in privacy test set the accuracy decreases. On the other hand, compared with the result of model 3, model 5 has about 3.58\% and 1.58\% improvement, which is brought by Horizontal Voting method. 

Serious overfitting is also observed in the model. The difference of accuracy between public and private test set is 0.0162. So combining Vertical Voting and Horizontal Voting is not a appropriate strategy, besides it costs much more computation resources.

\begin{table*}[htbp]
\caption{Classification accuracy of model 1-6. The best scores are in bold.}
\label{m5}
\vskip 0.15in
\begin{center}
\begin{small}
\begin{sc}
\begin{tabular}{l|llllll}
    \hline
    ~                           & Model 1 & model 2   & model 3   & model 4	& model 5  & model 6 \\ \hline
    Accuracy(public test set) & 0.55460 & 0.66660   & 0.65920   & 0.68220	& 0.68280 & \textbf{0.68540}\\
    Accuracy(private test set)& 0.54680 & 0.65120   & 0.65620   & 0.67240	& 0.66660 & \textbf{0.67440}\\ \hline
    \end{tabular}
\end{sc}
\end{small}
\end{center}
\vskip -0.1in
\end{table*}

\subsection{Model 6}
Sergey Yurgenson suggests that his non-linear ensemble method achieved a 10\% improvement in his shallow network. In Model 6, the random forest on 200 softmax output of deep neural network achieves accuracy of 0.68540 on public test set, and 0.67440 on private test set. The improvement is 2.82\% and 3.56\% compared to Model 2. and the method performs even better than model 4. Similar results have been observed in our other networks using this method, which indicates that this  ensemble method really learns something from probability output of neural network and adjusts to a better output.

\section{Conclusion and Future Work}
We focus on the classification problem without prior knowledge to the data. Using very limited number of labeled data and massive unlabeled data, we have achieved a good performance in ICML 2013 Black Box Leaning Challenge, by exploiting the power of deep neural networks.
 
In this work, we propose Vertical Voting, Horizontal Voting and Horizontal Stacked Ensemble method for deep neural network and test performance of them. We find that hierarchical representation in different layers may not lead to a better classification accuracy as expected. On the other hand, for representations in horizontal, both linear Horizontal Voting and Horizontal Stacked Ensemble methods can robustly improve the performance.

If we were provided with more knowledge about the data or more labeled training sets, we could have done more investigations and harvested deeper understanding for the representations in hierarchy. This exploration may be done on other datasets in the future. 

\section*{Acknowledgments} 
This research was supported by “the Fundamental Research Funds for the Central Universities” (Grant No. 2012RC0129), Important National Science \& Technology Specific Projects (Grant No. 2011ZX03002-005-01), National Natural Science Foundation of China (Grant No.61273217), and 111 Project of China under Grant No. B08004. Also we thank Ian Goodfellow's generous help in pylearn2-dev group. Finally we thank all participates in kaggle forum. You share ideas kindly, which will help everyone in the future.

\nocite{langley00}

\bibliography{example_paper}

\begin{thebibliography}{14}
\providecommand{\natexlab}[1]{#1}
\providecommand{\url}[1]{\texttt{#1}}
\expandafter\ifx\csname urlstyle\endcsname\relax
  \providecommand{\doi}[1]{doi: #1}\else
  \providecommand{\doi}{doi: \begingroup \urlstyle{rm}\Url}\fi

\bibitem[Bengio(2009)]{bengio2009learning}
Bengio, Yoshua.
\newblock Learning deep architectures for ai.
\newblock \emph{Foundations and Trends{\textregistered} in Machine Learning},
  2\penalty0 (1):\penalty0 1--127, 2009.

\bibitem[Bengio et~al.(2007)Bengio, Lamblin, Popovici, and
  Larochelle]{bengio2007greedy}
Bengio, Yoshua, Lamblin, Pascal, Popovici, Dan, and Larochelle, Hugo.
\newblock Greedy layer-wise training of deep networks.
\newblock \emph{Advances in neural information processing systems},
  19:\penalty0 153, 2007.

\bibitem[Bengio et~al.(2012)Bengio, Courville, and
  Vincent]{bengio2012representation}
Bengio, Yoshua, Courville, Aaron, and Vincent, Pascal.
\newblock Representation learning: A review and new perspectives.
\newblock \emph{arXiv preprint arXiv:1206.5538}, 2012.

\bibitem[Bergstra et~al.(2010)Bergstra, Breuleux, Bastien, Lamblin, Pascanu,
  Desjardins, Turian, Warde-Farley, and Bengio]{bergstra2010theano}
Bergstra, James, Breuleux, Olivier, Bastien, Fr{\'e}d{\'e}ric, Lamblin, Pascal,
  Pascanu, Razvan, Desjardins, Guillaume, Turian, Joseph, Warde-Farley, David,
  and Bengio, Yoshua.
\newblock Theano: a cpu and gpu math expression compiler.
\newblock In \emph{Proceedings of the Python for Scientific Computing
  Conference (SciPy)}, volume~4, 2010.

\bibitem[Bourlard \& Kamp(1988)Bourlard and Kamp]{bourlard1988auto}
Bourlard, Herv{\'e} and Kamp, Yves.
\newblock Auto-association by multilayer perceptrons and singular value
  decomposition.
\newblock \emph{Biological cybernetics}, 59\penalty0 (4-5):\penalty0 291--294,
  1988.

\bibitem[Erhan et~al.(2009)Erhan, Manzagol, Bengio, Bengio, and
  Vincent]{erhan2009difficulty}
Erhan, Dumitru, Manzagol, Pierre-Antoine, Bengio, Yoshua, Bengio, Samy, and
  Vincent, Pascal.
\newblock The difficulty of training deep architectures and the effect of
  unsupervised pre-training.
\newblock In \emph{Proceedings of The Twelfth International Conference on
  Artificial Intelligence and Statistics (AISTATS��09)}, pp.\  153--160.
  Citeseer, 2009.

\bibitem[Goodfellow et~al.(2013)Goodfellow, Warde-Farley, Mirza, Courville, and
  Bengio]{goodfellow2013maxout}
Goodfellow, Ian~J, Warde-Farley, David, Mirza, Mehdi, Courville, Aaron, and
  Bengio, Yoshua.
\newblock Maxout networks.
\newblock \emph{arXiv preprint arXiv:1302.4389v3}, 2013.

\bibitem[Hinton \& Zemel(1994)Hinton and Zemel]{hinton1994autoencoders}
Hinton, Geoffrey~E and Zemel, Richard~S.
\newblock Autoencoders, minimum description length, and helmholtz free energy.
\newblock \emph{Advances in neural information processing systems}, pp.\  3--3,
  1994.

\bibitem[Hinton et~al.(2006)Hinton, Osindero, and Teh]{hinton2006fast}
Hinton, Geoffrey~E, Osindero, Simon, and Teh, Yee-Whye.
\newblock A fast learning algorithm for deep belief nets.
\newblock \emph{Neural computation}, 18\penalty0 (7):\penalty0 1527--1554,
  2006.

\bibitem[Hinton et~al.(2012)Hinton, Srivastava, Krizhevsky, Sutskever, and
  Salakhutdinov]{hinton2012improving}
Hinton, Geoffrey~E, Srivastava, Nitish, Krizhevsky, Alex, Sutskever, Ilya, and
  Salakhutdinov, Ruslan~R.
\newblock Improving neural networks by preventing co-adaptation of feature
  detectors.
\newblock \emph{arXiv preprint arXiv:1207.0580}, 2012.

\bibitem[Le~Cun(1987)]{le1987modeles}
Le~Cun, Yann.
\newblock \emph{Mod{\`e}les connexionnistes de l'apprentissage}.
\newblock PhD thesis, 1987.

\bibitem[Pedregosa et~al.(2011)Pedregosa, Varoquaux, Gramfort, Michel, Thirion,
  Grisel, Blondel, Prettenhofer, Weiss, Dubourg, Vanderplas, Passos,
  Cournapeau, Brucher, Perrot, and Duchesnay]{scikit-learn}
Pedregosa, F., Varoquaux, G., Gramfort, A., Michel, V., Thirion, B., Grisel,
  O., Blondel, M., Prettenhofer, P., Weiss, R., Dubourg, V., Vanderplas, J.,
  Passos, A., Cournapeau, D., Brucher, M., Perrot, M., and Duchesnay, E.
\newblock Scikit-learn: Machine learning in {P}ython.
\newblock \emph{Journal of Machine Learning Research}, 12:\penalty0 2825--2830,
  2011.

\bibitem[Smolensky(1986)]{smolensky1986information}
Smolensky, Paul.
\newblock Information processing in dynamical systems: Foundations of harmony
  theory.
\newblock 1986.

\bibitem[Warde-Farley et~al.(2011)Warde-Farley, Goodfellow, Lamblin,
  Desjardins, Bastien, and Bengio]{Pylearn2}
Warde-Farley, David, Goodfellow, Ian, Lamblin, Pascal, Desjardins, Guillaume,
  Bastien, Fr{\'{e}}d{\'{e}}ric, and Bengio, Yoshua.
\newblock pylearn2, 2011.
\newblock \url{http://deeplearning.net/software/pylearn2}.

\end{thebibliography}
\bibliographystyle{icml2013}

\end{document}